\documentclass[lettersize,journal]{IEEEtran}
\usepackage{amsmath,amsfonts}
\usepackage{algorithmic}
\usepackage{algorithm}
\usepackage{array}
\usepackage[caption=false,font=normalsize,labelfont=sf,textfont=sf]{subfig}
\usepackage{textcomp}
\usepackage{stfloats}
\usepackage{url}
\usepackage{verbatim}
\usepackage{graphicx}
\usepackage{cite}
\usepackage{multirow}
\usepackage{colortbl}
\usepackage{xcolor}
\definecolor{best}{RGB}{255,165,165}
\definecolor{second}{RGB}{255,215,170}   
\definecolor{third}{RGB}{255,245,200}

\begin{document}

\title{GaussianSwap: Animatable Video Face Swapping with \\  3D Gaussian Splatting}

\author{Xuan Cheng, Jiahao Rao, Chengyang Li, Wenhao Wang, Weilin Chen, Lvqing Yang
\thanks{
Xuan Cheng, Jiahao Rao, Chengyang Li, Wenhao Wang, Weilin Chen and Lvqing Yang are with the School of Informatics, Xiamen University, Xiamen 361005, China.
\emph{Corresponding author: Lvqing Yang}.}
}

\markboth{Journal of \LaTeX\ Class Files,~Vol.~14, No.~8, August~2021}%
{Shell \MakeLowercase{\textit{et al.}}: A Sample Article Using IEEEtran.cls for IEEE Journals}


\maketitle

\begin{abstract}
We introduce \textbf{\textit{GaussianSwap}}, a novel video face swapping framework that constructs a 3D Gaussian Splatting based face avatar from a target video while transferring identity from a source image to the avatar. Conventional video swapping frameworks are limited to generating facial representations in pixel-based formats. The resulting swapped faces exist merely as a set of unstructured pixels without any capacity for animation or interactive manipulation. Our work introduces a paradigm shift from conventional pixel-based video generation to the creation of high-fidelity avatar with swapped faces. The framework first preprocesses target video to extract FLAME parameters, camera poses and segmentation masks, and then rigs 3D Gaussian splats to the FLAME model across frames, enabling dynamic facial control. To ensure identity preserving, we propose an compound identity  embedding constructed from three state-of-the-art face recognition models for avatar finetuning. Finally, we render the face-swapped avatar on the background frames to obtain the face-swapped video. Experimental results demonstrate that GaussianSwap achieves superior identity preservation, visual clarity and temporal consistency, while enabling previously unattainable interactive applications.
\end{abstract}

\begin{IEEEkeywords}
video face-swapping, 3DGS, face avatar
\end{IEEEkeywords} 

\section{Introduction}
Face swapping is a technique that transfers identity characteristics from a source image to a subject in a target image or video while preserving the target's non-identity attributes, including pose, facial expressions and background. This technique holds significant potential for applications in movie production, digital human and privacy protection. Since Deepfakes~\cite{mirsky2021deepfakesurvey} first emerged in 2017 and captured widespread attention, the field has maintained sustained research interest and continues to evolve.

Early methods in face swapping primarily concentrated on still images, such as GAN-based methods~\cite{nirkin2019fsgan, chen2020simswap, li2020faceshifter, wang2021hififace, zhu2021MegaFS} and Diffusion-based methods~\cite{kim2025diffface, zhao2023diffswap, liu2024DiffSFSR, baliah2025REFace}. To process a video, recent methods like DynamicFace~\cite{DynamicFace2025}, VividFace~\cite{VividFace2024} and HiFiVFS~\cite{HiFiVFS2024} begin to incorporate temporal attention mechanismm~\cite{Attention2017} to maintain temporal continuity across frames. However, the feature smoothing between frames often compromises identity preservation in favor of temporal stability. Achieving an optimal balance between attribute preservation and temporal coherence remains a critical issue in this field.

\begin{figure}
\centering
\includegraphics[width=0.45\textwidth]{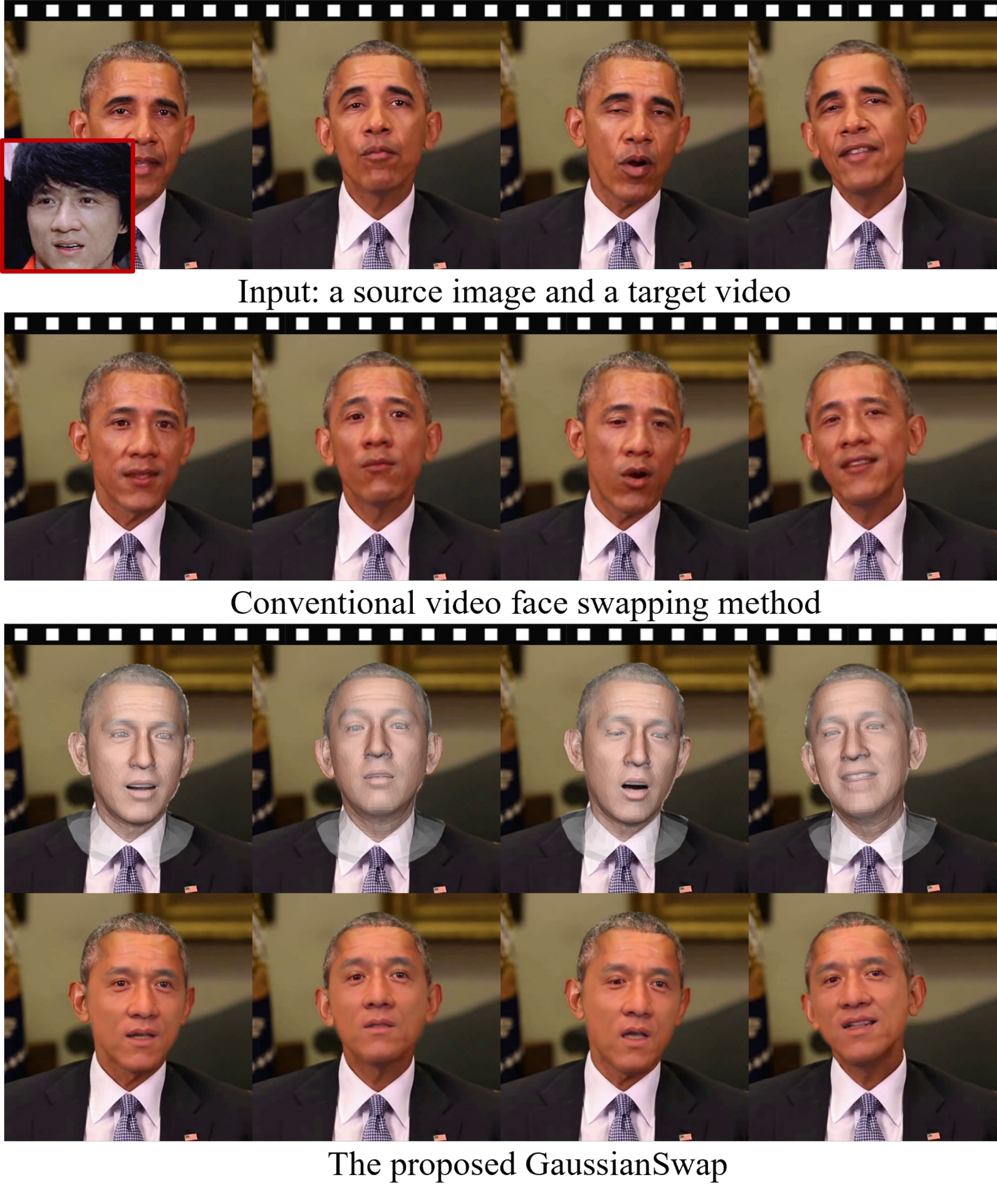} 
\caption{For video face swapping task, our GaussianSwap can generate not only face-swapped video (4th row) like conventional methods (2nd row) but also face-swapped avatar (3rd row), which can facilitate many interactive applications.}
\label{fig:teaser}
\end{figure}

Moreover, both the image and video face swapping methods are fundamentally limited to generating facial representations in pixel-based formats, namely images and videos. The resulting swapped faces exist merely as a set of unstructured pixels without any capacity for animation or interactive manipulation. The absence of parametric 3D facial representation in these pixel-based outputs means they can't be naturally integrated into interactive applications that require dynamic facial control, such as real-time expression editing, novel view rendering or responsive interaction in virtual environments. 

To address the aforementioned challenges in video face swapping, we propose a paradigm shift from conventional pixel-based video generation to the creation of high-fidelity avatars with swapped faces. We show an example in Fig. \ref{fig:teaser}. These head avatars retain all the functionalities of traditional methods when rendered to image plane, producing face-swapped videos of competitive temporal stability and visual quality. 
Crucially, because the face swapping is performed on the avatars constructed from the target video frames, temporal coherence is inherently preserved.
Moreover, the inherent animatability of these avatars unlocks a range of previously unattainable interactive applications, including video face reenactment, speech/text-driven facial animation and dynamic background manipulation in video conferencing. These capabilities can't be achieved by existing video face swapping methods without extensive, application-specific post-processing pipelines.

To this end, we propose GaussianSwap, a novel video face swapping framework that constructs a 3D Gaussian Splatting (3DGS) \cite{kerbl20233dgs} based head avatar from a monocular target video while preserving the identity characteristics extracted from a source image. Our framework employ 3DGS as the primary 3D head representation due to its unique combination of real-time rendering and high visual fidelity, which enables seamless avatar-environment interactions. The pipeline of GaussianSwap begins by preprocessing the target video through 3D face tracking and head-torso segmentation, extracting per-frame FLAME \cite{li2017FLAME} parameters, camera poses and segmentation masks. We then construct an animatable 3DGS-based avatar by rigging the Gaussian splats to the FLAME parametric face model across all frames. To ensure identity fidelity between the avatar and the source image, we further finetune the avatar through incorporating a compound identity  embedding, which is constructed by three state-of-the-art face recognition models. The constructed head avatar can be used not only to generate face-swapped videos, but also to seamlessly integrate into previously unattainable interactive applications.

The contributions of this paper are as follows:
\begin{itemize}
    \item  We propose a novel video swapping framework that constructs a high-fidelity 3DGS-based head avatar from a target video while transferring identity from a source image to the avatar.
    
    \item We propose a novel identity preserving approach in 3DGS optimization that uses compound identity embedding to comprehensively capture identity characteristic. 
    
    \item We show that the face swapping results can be seamlessly used in several new application scenarios: video face reenactment, speech-driven facial animation and dynamic background manipulation.
\end{itemize}

\section{Related Work}
\subsection{Image Face Swapping}
Image face swapping methods can generally be categorized into two groups based on their generative models: GAN-based methods and Diffusion-based methods.

FSGAN~\cite{nirkin2019fsgan} pioneered the integration of GANs with face swapping, introducing a multi-network framework capable of simultaneous face swapping and reenactment. Subsequently, SimSwap~\cite{chen2020simswap} proposed an identity injection module for feature-level identity transfer and a weak feature matching loss to implicitly preserve facial attributes, enabling identity-agnostic face swapping. In the same year, FaceShifter~\cite{li2020faceshifter} introduced a two-stage framework for high-quality face swapping under occlusions. Unlike methods relying solely on face recognition networks for identity preservation, HifiFace~\cite{wang2021hififace} leveraged 3DMM-based~\cite{blanz2023morphable} facial geometry to enhance identity fidelity through shape control. Additionally, advancements in StyleGAN~\cite{karras2019stylegan1, karras2020stylegan2, EMEF2023} significantly propelled face-swapping techniques. For instance, MegaFS~\cite{zhu2021MegaFS} utilized its generator to achieve megapixel-resolution single-shot face swapping.

Driven by Diffusion model's generative prowess and training stability, researchers have recently explored diffusion-based face-swapping methods. DiffFace~\cite{kim2025diffface} first employed a Diffusion model for face swapping, training an identity-conditioned DDPM~\cite{ho2020diffusion1} and incorporating a facial expert during sampling to transfer identities while preserving target attributes. Building on this, DiffSwap~\cite{zhao2023diffswap} guided diffusion using identity, landmarks and facial attributes, redefining face swapping as conditional image inpainting. To jointly handle face swapping and reenactment, DiffSFSR~\cite{liu2024DiffSFSR} achieved fine-grained identity and expression control via diffusion. Most recently, REFace~\cite{baliah2025REFace} enhanced identity transfer and attribute preservation using CLIP~\cite{radford2021CLIP} features while accelerating high-quality generation through simplified denoising.

\subsection{Video Face Swapping}
While image face swapping methods can process a video by performing frame-by-frame swapping, they can't maintain temporal consistency. Ghost~\cite{groshev2022ghost} mitigated the face jittering between adjacent frames through landmark smoothing. Recent advances in video diffusion models~\cite{SVD2023} have led to methods like DynamicFace~\cite{DynamicFace2025}, VividFace~\cite{VividFace2024} and HiFiVFS~\cite{HiFiVFS2024}, which employed temporal attention mechanism~\cite{Attention2017} to enhance temporal consistency.
Considering these video Diffusion-based methods often require substantial computational resources, CanonSwap~\cite{CanonSwap2025} resolves temporal instability by decoupling pose variations from identity transfer in canonical space.

Although these video face swapping methods can produce stable temporally consistent results, they remain limited to unstructured pixel representations. In contrast, our proposed GaussianSwap simultaneously achieves high-quality video face swapping comparable to state-of-the-art methods, and generation of dynamically controllable face avatar with inherent 3D structure.

\begin{figure*}
\centering
\includegraphics[width=1.0\textwidth]{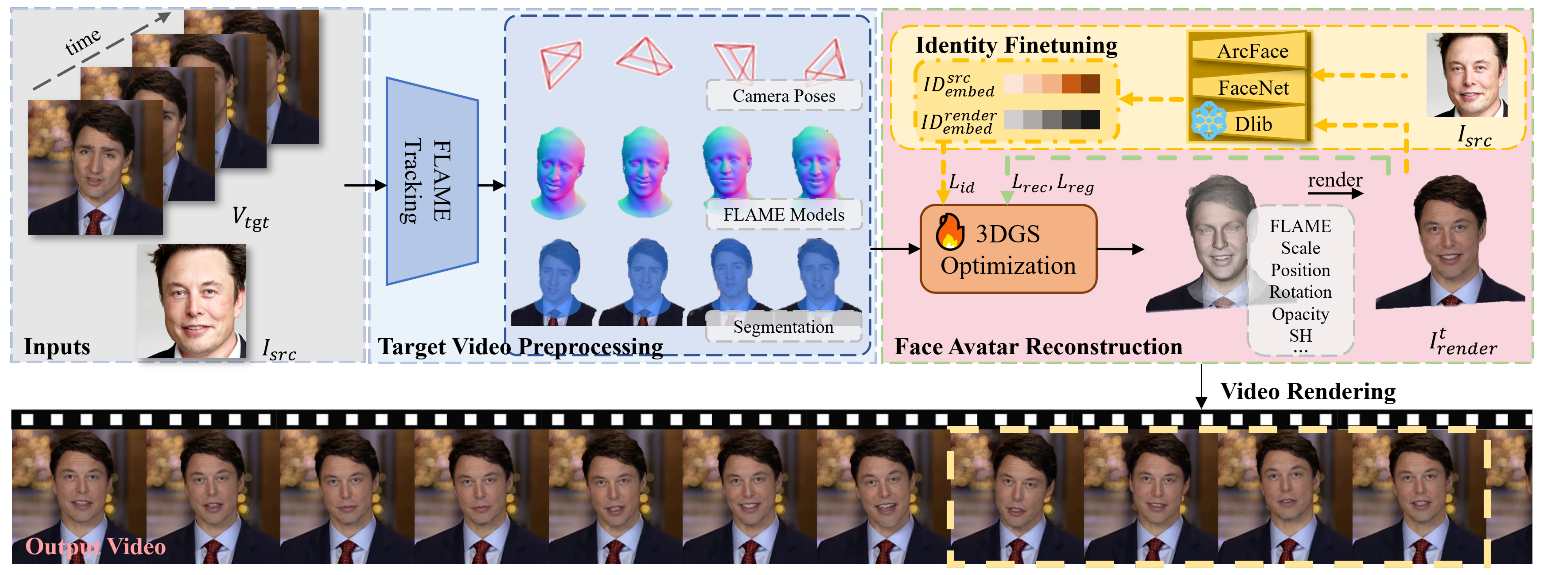} 
\caption{\textbf{Overview of the GaussianSwap framework.} 
The framework takes a source image $I_{src}$ and a target video $V_{tgt}$ as input, and generates a high-fidelity avatar from $V_{tgt}$ with the face swapped to match $I_{src}$. In the pipeline, FLAME tracking is first performed on the video sequence $V_{tgt}$ to obtain per-frame FLAME parameters, camera poses and segmentation/matting masks. A 3DGS-based face avatar is then built using the FLAME tracking data, where the 3D Gaussians are dynamically bound to the triangular faces of the FLAME mesh models through 3DGS optimization. To enforce the identity similarity between the avatar and $I_{src}$, the avatar undergoes additional training iterations supervised by three SOTA face recognition models: ArcFace, FaceNet and Dlib. Finally, the high-fidelity, face-swapped avatar is generated, which can be further rendered into the face-swapped video.}
\label{fig:overview}
\end{figure*}

\subsection{3D Head Avatar Reconstruction}
While 3D face reconstruction \cite{blanz2023morphable, li2017FLAME, DECA2021, DCT2024, FaceRefiner2024, DNPM} focuses on the facial region, 3D head avatar reconstruction demands a comprehensive model of the entire head geometry and its appearance.
Gafni et al.~\cite{gafni2021dynamic} introduced the first 4D head reconstruction method using Neural Radiance Fields~\cite{mildenhall2020nerf}, enabling expression-driven control through additional expression conditions in the trained model. The following year, ~\cite{grassal2022neural} enhanced neural representations with an explicit articulated head model, ensuring high-fidelity reconstruction under significant viewpoint changes. Concurrently, IMAvatar~\cite{zheng2022iamavatar} integrated blendshapes and skinning into volumetric rendering to manipulate facial expressions and pose deformations. Departing from earlier methods that relied on 3DMM expression parameters for driving, LatentAvatar~\cite{xu2023latentavatar} proposed an implicit expression encoding approach. INSTA~\cite{zielonka2023instant} leveraged InstantNGP's~\cite{muller2022instantNGP} rapid rendering, reducing training time to 10 minutes while preserving reconstruction quality.

Despite the advantages of implicit modeling, researchers continue exploring more practical explicit approaches. Beginning with PointAvatar~\cite{zheng2023pointavatar} which used point cloud representations, the emergence of 3DGS~\cite{kerbl20233dgs} has spurred numerous works adopting explicit modeling schemes~\cite{giebenhain2024npga, shao2024splattingavatar, xiang2024flashavatar, saito2024relightable}. Capitalizing on 3DGS's efficiency, SplattingAvatar~\cite{shao2024splattingavatar} achieved real-time rendering at 300 FPS in GPU and 30 FPS in mobile device. GaussianAvatars~\cite{qian2024gaussianavatars} bounds 3D Gaussian ellipsoids to FLAME~\cite{li2017FLAME} meshes during training, enabling avatar driving through new FLAME inputs. NPGA~\cite{giebenhain2024npga} integrated MonoNPHM's~\cite{giebenhain2024mononphm} expression priors for finer-grained control, while RelightableGaussian~\cite{saito2024relightable} employed detailed appearance modeling (diffuse color, specular highlights and normals) to achieve relightable rendering. Learn2Talk~\cite{Learn2Talk2025} and TalkingEyes~\cite{TalkingEyes2025} used speech signals to drive a pre-bulit 3DGS-based head avatar. 

\section{framework}
\label{sect:framework}
To reconstruct a high-quality and controllable swapped dynamic face avatar, GaussianSwap takes a monocular target video $V_{tgt}$ and a source face image $I_{src}$ as input. The pipeline of GaussianSwap is illustrated in Fig. \ref{fig:overview}, which contains four main steps: target video preprocessing, face avatar reconstruction, identity finetuning and face-swapped video generation.

\subsection{Target Video Preprocessing}
Given a monocular target video, the video preprocessing step extracts the high-quality 3D face tracking data from it, thus enabling the creation of animatable face avatar in the next step.

The preprocessing step firstly applies Robust Video Matting~\cite{lin2022robust} on each frame of the video to separate the subject (primarily the head and torso) from the background. Next, the FLAME~\cite{li2017FLAME} tracking
produces optimized FLAME parameters that accurately represent the subject's facial geometry and appearance throughout the video sequence. These optimized FLAME parameters include shape, expression, pose and skinning weights. Similar to DECA \cite{DECA2021}, the FLAME optimization employs a loss function comprising facial landmark loss, image reconstruction loss and FLAME parameter regularization.

DECA performs FLAME optimization on only a single image. When applied to video FLAME tracking, it fails to maintain consistency in FLAME parameters across frames. To address this issue, we first conduct frame-by-frame FLAME optimization, then randomly sample frames from the sequence to jointly optimize their FLAME parameters for improved temporal consistency.

\subsection{Face Avatar Reconstruction}
We use a binding method similar to GaussianAvatars~\cite{qian2024gaussianavatars} to associate 3D Gaussians with the tracked FLAME mesh models. 
The 3D Gaussian is initialized to each triangular face in the local coordinate system by setting the center $\mu$ to the origin, the rotation $r$ to the identity rotation matrix and the scale $s$ is to the unit vector. Then, these 3D Gaussians are moved with their parent faces across time steps. Specifically, the 3D Gaussian is transformed from the local space to the global space by:
\begin{equation}
\begin{aligned}
\label{local_to_world}
\tilde{r} & = Kr \\
\tilde{\mu} & = lK\mu + V \\
\tilde{s} & = ls,
\end{aligned} 
\end{equation}
where $\tilde{\mu}$, and $\tilde{s}$ denote the center, rotation and scale in the global space, $K$ and $V$ denote the rotation and translation transformation matrices computed from each face, and scalar $l$ represents the size of triangular face. 
This adaptive scaling $l$ ensures proportional Gaussian dimensions: larger faces maintain association with larger Gaussians and vice versa.

The loss function for optimizing the 3DGS-based face avatar against the target video comprises a reconstruction loss and a regularization loss. 

\textbf{Reconstruction Loss} enforces photometric consistency between the avatar-rendered images $I_{render}^{t}$ and the corresponding target video frames $I_{tgt}^{t}$. We adopt the $L_1$ loss and the SSIM loss, which have been widely used in various 3DGS-based works:
\begin{equation}
\begin{aligned}
\label{recon_loss}
L_{rec} = (1 - \lambda_{ssim}) \|I_{render}^{t} - I_{tgt}^{t} \|_{1} + \\ \lambda_{ssim}SSIM(I_{render}^{t}, I_{tgt}^{t}). 
\end{aligned}
\end{equation}
The hyperparameter $\lambda_{ssim}$ that balances the two losses is set to 0.2. 

\textbf{Regularization Loss} regularizes the center $\mu$ and the scale $s$ of 3D Gaussian. Although the reconstruction loss can generate high-quality rendered images, the lack of additional constraints may lead to poor alignment between 3D Gaussians and triangular faces. Therefore, we apply regularization to the center $\mu$ and the scale $s$ of each 3D Gaussian to improve alignment quality. The regularization loss $L_{reg}$ consists of two components: scale loss $L_{scale}$ and position loss $L_{pos}$, which is defined as:
\begin{equation}
\label{reg_loss}
L_{reg} = \lambda_{scale} L_{scale} + \lambda_{pos} L_{pos}.
\end{equation}
The hyperparameters $\lambda_{scale}$ and $\lambda_{pos}$ are set to 1 and 0.01 respectively.

The scale loss $L_{scale}$ is used to constrain the size of each 3D Gaussian, which is formulated as:
\begin{align}
\label{reg_scale}
L_{scale} = \|max(s, \phi_{scale})\|_{2}.
\end{align} 
$\phi_{scale}$ denotes the threshold, which is set to 0.6. $L_{scale}$ encourages the scale $\tilde{s}$ of a 3D Gaussian in the global space to be no larger than 0.6 times the size of its associated triangular face. This scaling limitation prevents small triangular surfaces from being assigned disproportionately large 3D Gaussians, which could otherwise introduce significant visual artifacts in avatar animation.

The position loss $L_{pos}$ maintains spatial coherence by constraining each 3D Gaussian to remain proximal to its associated triangular face. For example, a 3D Gaussian initially bound to the nose should maintain its spatial localization throughout optimization, preventing erroneous migration to the eyes. 
$L_{pos}$ is defined as:
\begin{align}
\label{reg_pos}
L_{pos} = \|max(\mu, \phi_{pos})\|_{2},
\end{align}
where the threshold $\phi_{pos}$ constrains the permissible deviation of each 3D Gaussian from its associated triangular face, and it is usually set to 1.0.

\subsection{Identity Finetuning}
After the face avatar reconstruction, we obtain the 3DGS-based face avatar built from the target video. To transfer the identity from source image to this avatar, we innovatively incorporate an identity loss based on the compound identity  embedding. The motivation of the compound identity  embedding is that a single identity embedding is usually biased and thus can't comprehensively capture the identity characteristic.
Three state-of-the-art face recognition models are selected to construct the compound identity  embedding, which includes Dlib ~\cite{king2009dlib}, FaceNet ~\cite{schroff2015facenet} and ArcFace ~\cite{deng2019arcface}. The identity loss is defined as:
\begin{equation}
\label{id_loss}
L_{id} = \sum_{k=1}^{K}\lambda_{k} \left ( 1-\cos\left ( E_{id}^{k}\left ( I_{src}  \right ),E_{id}^{k}\left ( I_{render}^t \right )    \right )    \right ) ,
\end{equation}
where $E_{id}^{k}\left ( \cdot \right )$ represents one of the three ($K=3$) identity encoder constructed by the pretrained face recognition models, $\lambda_{k}$ controls the contribution of each identity encoder, and $\cos(\cdot , \cdot)$ denotes the cosine similarity measuring the similarity of two identity embeddings.
In our experiments, we set $\lambda_{1}=0.9$, $\lambda_{2}=0.001$ and 
$\lambda_{3}=0.1$.

Note that the ArcFace model requires aligned and cropped face images as input. To avoid additional computational overhead during optimization, we precompute the necessary affine transformation matrices in the video preprocessing stage.

The overall loss function in the avatar finetuning stage is defined as the combination of all the aforementioned losses:
\begin{equation}
\label{total_loss}
L_{total} = L_{rec} + L_{reg} + \lambda_{id}L_{id},
\end{equation}
where the hyperparameter $\lambda_{id}$ is set to 0.1.

To obtain finer facial details, GaussianSwap adopts a different adaptive density control mechanism compared to the original 3DGS. It records the indices of the triangular faces to their associated 3D Gaussians, thereby ensuring that newly added Gaussians preserve the same binding relationships as the originals.

\subsection{Face-swapped Video Rendering}
To produce the final face-swapped video, the face-swapped avatar is frame-by-frame rendered on the background images in the target video. We develop a robust face video fusion method to seamlessly blend the avatar-rendered images $I_{swapped}$ with the background frames $I_{tgt}$.

Firstly, we conduct face parsing on $I_{swapped}$ to obtain the head mask $M_{swapped}$. Then, we conduct video matting~\cite{lin2022robust} on $I_{tgt}$ to obtain the foreground mask $M_{tgt}$. Either $M_{swapped}$ or $M_{tgt}$ alone could be used for blending. However, our experiments show that the combination of them can yield better blending results. Hence, we define the fused mask $M_{fuse}$ as:
\begin{equation}
\begin{aligned}
\label{get_fused_mask}
M_{fuse} &= M_{swapped} \cdot M_{tgt}. \\
\end{aligned}
\end{equation}
Due to the sharp boundaries and potential inaccuracies of $M_{fuse}$, directly using $M_{fuse}$ will introduce visual artifacts along the blending boundary between the background and the head. To address this issue, we also conduct edge erosion and Gaussian smoothing on $M_{fuse}$.

Finally, based on $M_{fuse}$, we perform a linear combination of $I_{swapped}$ and $I_{tgt}$ to obtain the final face-swapped image in each frame, which is defined as:
\begin{equation}
\label{add_bg}
I_{final} = I_{swapped} \cdot M_{fuse} + I_{tgt} \cdot (1 - M_{fuse}).
\end{equation}

\subsection{Implementation Details}
Most hyperparameters settings have already introduced in the above subsections, and remain fixed for all inputs across different subjects and datasets. We use Adam for the 3DGS parameter optimization with the same learning rates as 3DGS~\cite{kerbl20233dgs} implementation. We reduce the spherical harmonics order to one. The 3DGS-based face avatar is trained on a target video for totally 600,000 iterations, followed by an additional 120,000 iterations for identity finetuning.
Constructing a face-swapped face avatar on an NVIDIA RTX 4090 GPU requires approximately 6 to 10 hours, with duration varying according to the number of 3D Gaussians (larger quantities result in longer training times).

\section{Experiments}
\subsection{Quantitative Evaluation}
\subsubsection{Datasets}
We conduct quantitative evaluation on two datasets:  
FaceForensics++ (FF++)~\cite{2019FF++} and INSTA~\cite{zielonka2023instant}. 
FF++ is commonly used to evaluate face swapping performance in images/videos, while INSTA typically assesses head/face avatar reconstruction quality. From these datasets, we randomly select five monocular videos from FF++ and five from INSTA as the target videos, along with four portraits of globally recognized individuals as the source images. This creates 40 challenging target-source pairs covering same-gender, cross-gender and cross-ethnicity scenarios. Our GaussianSwap produces 20 face-swapped avatars on FF++ and 20 face-swapped avatars on INSTA.  
To ensure accurate foreground supervision during training, we use background-removed images as target frames.


\subsubsection{Evaluation Metrics}
We evaluate the face swapping performance using both the image-based face swapping metrics and the video consistency metrics. The image-related metrics include identity similarity (IDs), pose error (Pose), and expression error (Exp), while the video-related metrics include video identity distance (VIDD). When computing IDs, since ArcFace~\cite{deng2019arcface}, FaceNet~\cite{schroff2015facenet} and Dlib~\cite{king2009dlib} have been used in GaussianSwap's  training process, we employ BlendFace~\cite{reface2023} as an independent identity feature extractor to ensure unbiased evaluation. Higher IDs score indicates better identity preservation. When computing Exp and Pose, we use Deep3DFaceRecon~\cite{Deep3D2019} and HopeNet~\cite{HopeNet2020} to estimate expression and pose parameters for the target video and the swapped video respectively. We then calculate the mean Euclidean distance between corresponding parameters to quantify non-identity attributes preservation. For VIDD, we follow FOS~\cite{VIDD2024} to assess the temporal consistency between consecutive video frames. We average the evaluation metric values across all test data to obtain the overall performance score in each evaluation metric.

\begin{table}
\caption{Quantitative evaluation results on FF++ dataset. The best score in each metric is marked with bold. Red, orange, and yellow shading indicate \colorbox{best}{1st}, \colorbox{second}{2nd}, \colorbox{third}{3rd} place, respectively.}
\label{tab:FF++_comparative}
\renewcommand{\arraystretch}{1.1} 
\centering
\setlength{\tabcolsep}{3pt}
\begin{tabular}{l|ccc|c}
\hline
\multirow{2}{*}{Methods} & \multicolumn{3}{c|}{Image Quality} & \multicolumn{1}{c}{Video Quality} \\
\cline{2-5}
 & IDs$\uparrow$ & Pose$\downarrow$ & Exp$\downarrow$ & VIDD$\downarrow$  \\
\hline
SimSwap   & 55.2 & \cellcolor{second}\textbf{1.69} & \cellcolor{best}\textbf{2.41} & \cellcolor{best}\textbf{0.1477} \\
FaceDancer       & 45.9 & \cellcolor{second}\textbf{2.42} & 2.81 & 0.1525  \\
E4S        & 45.2 & 3.03 & 2.99 & 0.2035 \\
Ghost       & \cellcolor{second}\textbf{70.7} & 1.95 & 2.88  & \cellcolor{third}\textbf{0.1504} \\
CanonSwap       & \cellcolor{third}\textbf{62.1} & \cellcolor{best}\textbf{1.58} & 2.42  & 0.1510 \\
\hline
GaussianSwap           & \cellcolor{best}\textbf{71.7} & \cellcolor{third}\textbf{1.84} & \cellcolor{third}\textbf{2.51} & \cellcolor{second}\textbf{0.1472} \\
\hline
\end{tabular}
\end{table}

\begin{table}
\caption{Quantitative evaluation results on INSTA dataset. The best score in each metric is marked with bold. Red, orange, and yellow shading indicate \colorbox{best}{1st}, \colorbox{second}{2nd}, \colorbox{third}{3rd} place, respectively.}
\label{tab:INSTA_comparative}
\renewcommand{\arraystretch}{1.1} 
\centering
\setlength{\tabcolsep}{3pt}
\begin{tabular}{l|ccc|c}
\hline
\multirow{2}{*}{Methods} & \multicolumn{3}{c|}{Image Quality} & \multicolumn{1}{c}{Video Quality} \\
\cline{2-5}
 & IDs$\uparrow$ & Pose$\downarrow$ & Exp$\downarrow$ & VIDD$\downarrow$ \\
\hline

SimSwap   & 59.8 & \cellcolor{third}\textbf{2.20} & \cellcolor{second}\textbf{2.69} & \cellcolor{third}\textbf{0.3159}  \\
FaceDancer       & 43.3 & 2.39 & 2.81 & 0.3219 \\
E4S        & 51.0 & 3.22 & 3.21 & 0.4475 \\
Ghost       & \cellcolor{second}\textbf{78.1} & 2.79 & 3.06  & 0.3172 \\
CanonSwap       & \cellcolor{third}\textbf{64.8} & \cellcolor{best}\textbf{1.81} & \cellcolor{best}\textbf{2.54}  & \cellcolor{second}\textbf{0.3049} \\
\hline
GaussianSwap           & \cellcolor{best}\textbf{81.5} & \cellcolor{second}\textbf{2.06} & \cellcolor{third}\textbf{2.72} & \cellcolor{best}\textbf{0.2808}\\
\hline
\end{tabular}
\end{table}

\begin{figure*} 
\centering
\includegraphics[width=0.98\textwidth]{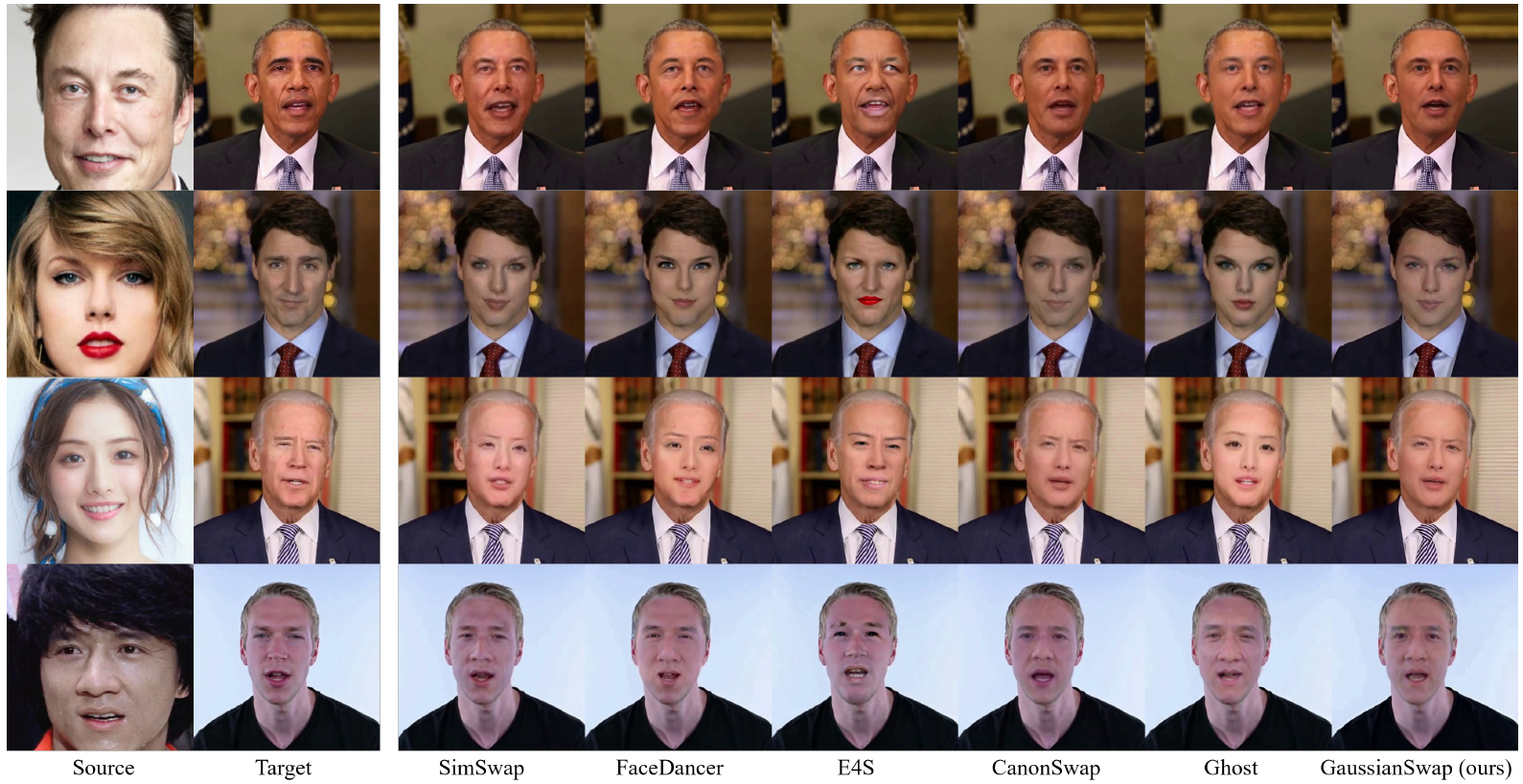} 
\caption{Qualitative results on INSTA. GaussianSwap achieves accurate identity transfer and  high-quality visual appearance in the swapped faces. More details are visible in the enlarged view.}
\label{fig:INSTA_compare}
\end{figure*}

\subsubsection{Competitors}
We compare our GaussianSwap with the image face swapping methods like SimSwap~\cite{chen2020simswap}, FaceDancer~\cite{FaceDancer2023} and E4S~\cite{E4S2023}, and the video face swapping methods like Ghost~\cite{groshev2022ghost} and CanonSwap~\cite{CanonSwap2025}. All the competing methods are open-source and we directly use their pre-traiend models for evaluation. Since the recent video face swapping methods including DynamicFace~\cite{DynamicFace2025}, 
and HiFiVFS~\cite{HiFiVFS2024} have not released their code or pre-trained models, we are unable to include them in the quantitative comparison.

\subsubsection{Results}
Tab. \ref{tab:FF++_comparative} and Tab. \ref{tab:INSTA_comparative} present the quantitative evaluation results on the FF++ and INSTA datasets, respectively. Our GaussianSwap demonstrates superior performance compared to competing methods, achieving significantly higher identity similarity scores on both datasets. These results confirm the effectiveness of our compound identity  embedding for identity preservation. Notably, GaussianSwap exhibits outstanding temporal consistency, achieving the higher video identity consistency scores on both FF++ and INSTA. By leveraging all frames from the target video in 3DGS optimization for avatar construction, GaussianSwap maintains exceptional temporal stability throughout the generated sequences. 
Furthermore, GaussianSwap maintains competitive performance in preserving non-identity attributes (pose and expression), achieving results comparable to leading methods while delivering superior identity preservation and temporal consistency.

\subsection{Qualitative Evaluation}
\begin{figure*} 
\centering
\includegraphics[width=0.98\textwidth]{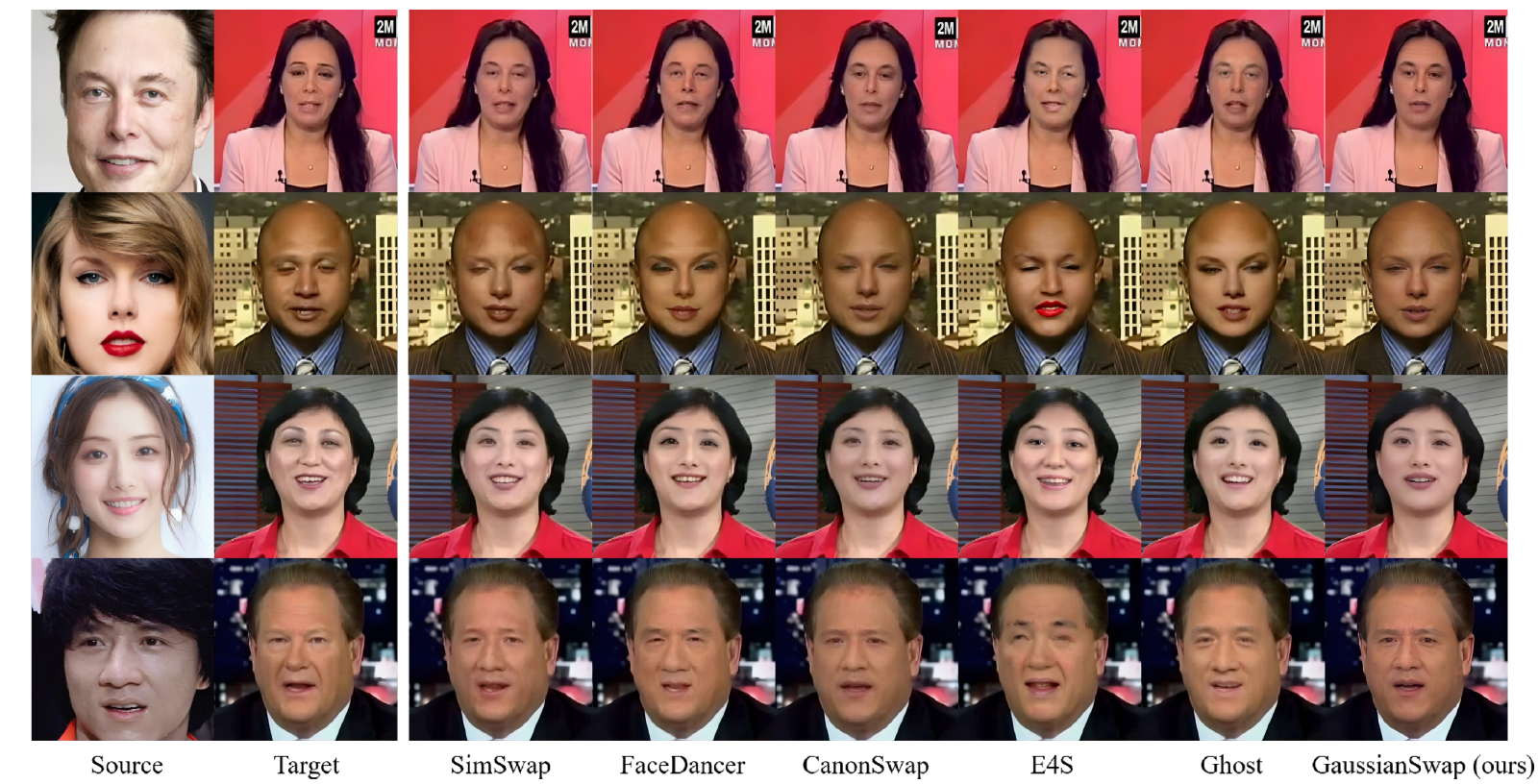} 
\caption{Qualitative results on FF++. GaussianSwap achieves accurate identity transfer and high-quality visual appearance in the swapped faces. More details are visible in the enlarged view.}
\label{fig:FF++_compare}
\end{figure*}

\begin{figure*} 
\centering
\includegraphics[width=0.98\textwidth]{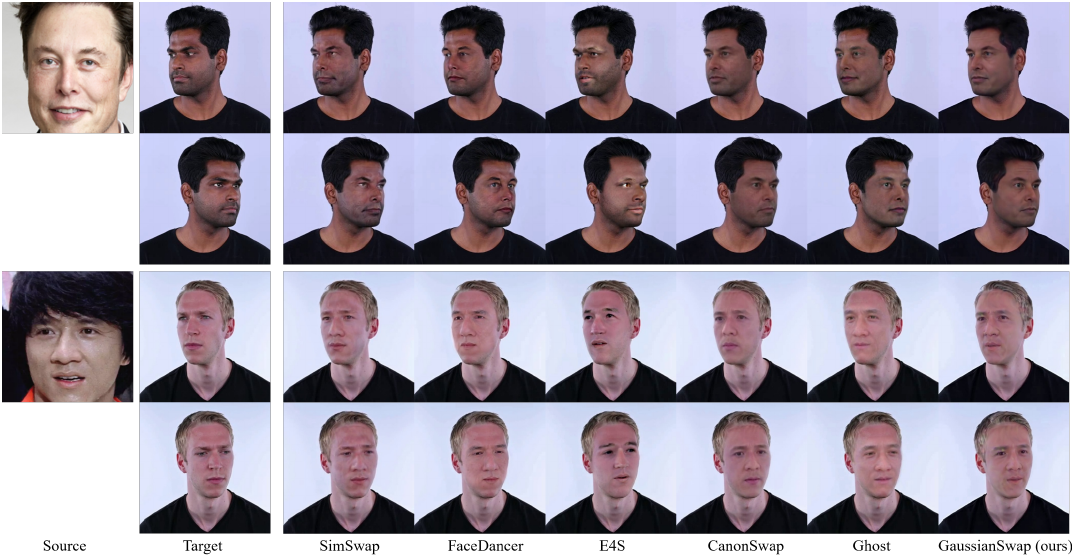} 
\caption{Face swapping results for side-view faces on INSTA. GaussianSwap performs effectively on these challenging poses. More details are visible in the enlarged view.}
\label{fig:INSTA_side}
\end{figure*}

Fig. \ref{fig:INSTA_compare} and Fig. \ref{fig:FF++_compare} show the qualitative comparisons on INSTA and FF++, respectively. The results show that our GaussianSwap not only achieves accurate identity transfer
but also presents high-quality visual appearance in the swapped faces (e.g., eyebrows, eyes, lips and teeth). Fig. \ref{fig:INSTA_side} shows the qualitative comparisons on side-view faces. GaussianSwap also performs effectively on these challenging poses. The visual comparisons on video face swapping are presented in the supplementary video.

\subsection{User Study}

For more comprehensive evaluation, we have conducted a perceptual user study to evaluate the video face swapping quality of our GaussianSwap and the five baseline methods, which include SimSwap~\cite{chen2020simswap}, FaceDancer~\cite{FaceDancer2023}, E4S~\cite{E4S2023} Ghost~\cite{groshev2022ghost}, and CanonSwap~\cite{CanonSwap2025}.
Specifically, we adopt the A/B test in a random order to compare our GaussianSwap with the baseline methods. In each comparison, the participants were asked to answer the three questions with A or B: 1) which resulting video has more similarity with the source image? 2) which resulting video has better temporal consistency? 3) which resulting video has better visual image quality? 
We randomly selected 18 videos from both INSTA and FF++, and used them to generate 18 A vs. B pairs. Some videos used in the user study have been presented in the supplementary video. 15 participants took part in the study, finally yielding 270 entries. We calculated the ratio of participants who prefer our method over the baseline in each evaluation metric. The percentage of A/B testing is tabulated in Tab. \ref{tab:userstudy}, which shows that participants consistently preferred our method over all the five baseline methods in terms of identity preservation, temporal consistency and visual image quality.

\begin{table} [h]
\caption{User study results. The percentage of answers where our method is preferred over the baseline method is listed.} 
\label{tab:userstudy}
\centering
\fontsize{8}{10}\selectfont
\begin{tabular}{l|ccc}
	\hline
    {Ours vs. Baseline}  & Identity & Temporal & Visual\cr
	\hline
    Ours vs. SimSwap & 76$\%$ & 56$\%$ & 82$\%$\cr
    Ours vs. FaceDancer & 89$\%$ & 67$\%$ & 73$\%$\cr
    Ours vs. E4S & 100$\%$ & 84$\%$ & 93$\%$ \cr
    Ours vs. Ghost & 58$\%$ & 65$\%$ & 76$\%$\cr
    Ours vs. CanonSwap &78$\%$ &69$\%$ &91$\%$ \cr
	\hline
\end{tabular}
\end{table}

\begin{table}
\caption{Quantitative evaluation results in ablation study. The best score in each metric is marked with bold. Red, orange, and yellow shading indicate \colorbox{best}{1st}, \colorbox{second}{2nd}, \colorbox{third}{3rd} place, respectively.}
\label{tab:ablation}
\renewcommand{\arraystretch}{1.1}
\centering
\setlength{\tabcolsep}{3pt}
\begin{tabular}{l@{\hskip 15pt}|cccc}
\hline
Methods & IDs$\uparrow$ & Exp$\downarrow$ & Pose$\downarrow$ & VIDD$\downarrow$ \\
\hline
w/o \textbf{a}           & 56.59 & \cellcolor{third}\textbf{2.60} & 2.17  & \cellcolor{best}\textbf{0.2692} \\
w/o \textbf{f}           & \cellcolor{second}\textbf{81.98} & \cellcolor{second}\textbf{2.58} & \cellcolor{best}\textbf{1.68}  & 0.2896 \\
w/o \textbf{d}           & \cellcolor{third}\textbf{81.34} & 2.63 & \cellcolor{third}\textbf{1.90}  & \cellcolor{third}\textbf{0.2751} \\
\hline
full model      & \cellcolor{best}\textbf{82.13} & \cellcolor{best}\textbf{2.57} & \cellcolor{second}\textbf{1.73} & \cellcolor{second}\textbf{0.2746}  \\
\hline
\end{tabular}
\end{table}

\subsection{Ablation Study}
To demonstrate the effectiveness of the compound identity embedding, we sequentially remove each identity embedding when computing the identity loss. The three ablation methods include: 1) w/o \textbf{a}, removing ArcFace model; 2) w/o \textbf{f}, removing FaceNet model; 3) w/o \textbf{d}, removing Dlib model.

The ablation study is conducted on INSTA. The quantitative evaluation results are reported in Tab. \ref{tab:ablation}. Compared to the full model, the removal of any individual embedding results in significant degradation of identity similarity performance. Notably, the absence of the ArcFace embedding causes the most substantial performance drop, confirming its critical role in identity preservation. To conclude, this quantitative ablation study confirms the complementary effectiveness of the three identity embeddings, and validates the necessity of
integrating all of them to achieve optimal video face swapping results.

\begin{figure}
\centering
\includegraphics[width=0.48\textwidth]{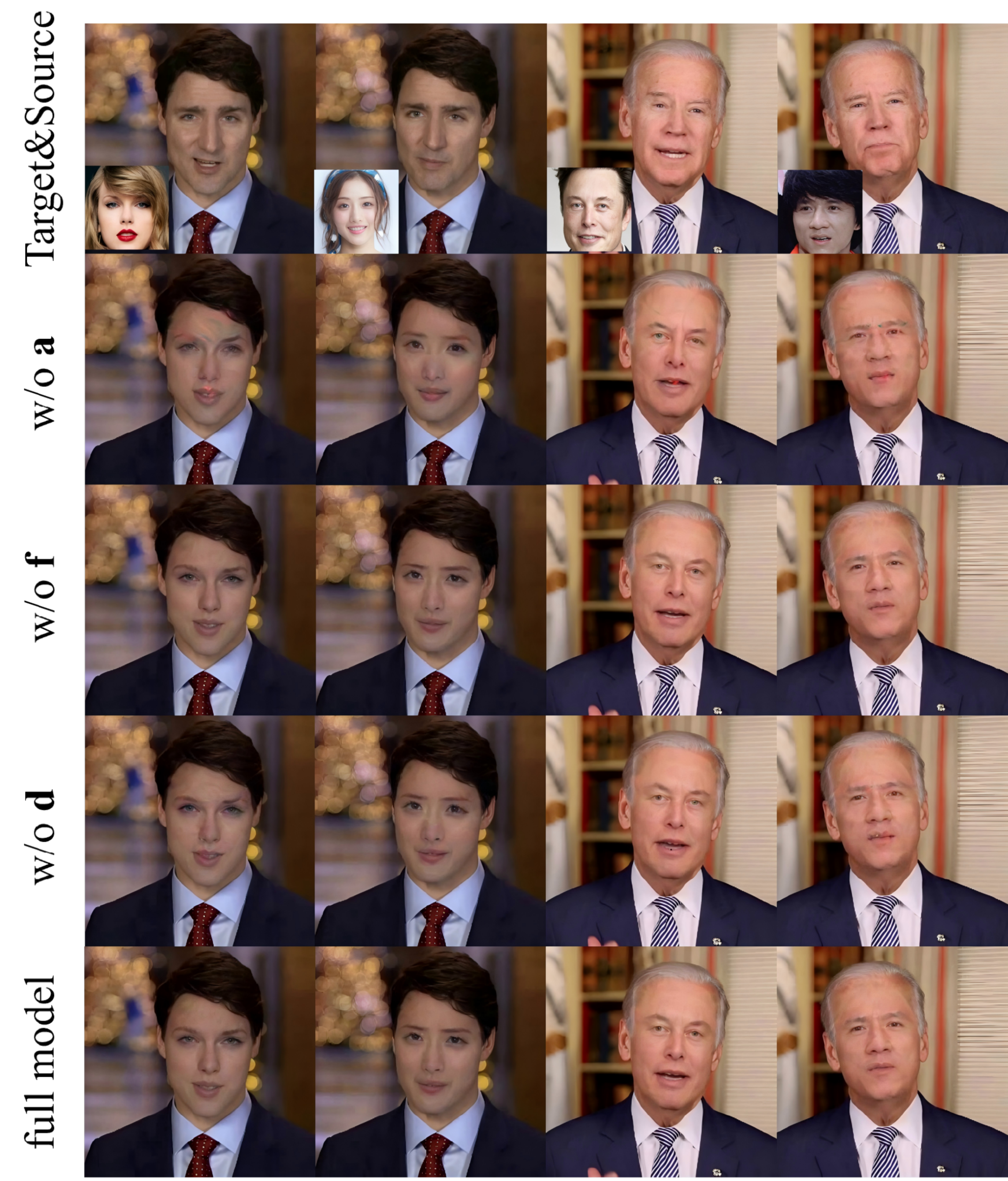} 
\caption{Qualitative results in ablation study. Removing any individual identity embedding
results in the degradation of image quality. }
\label{fig:ablation}
\end{figure}

Fig.~\ref{fig:ablation} shows visual comparisons. As shown in the 2nd row, removing ArcFace leads to obvious artifacts, color distortions and inaccurate identity transfer. The 3rd row  demonstrates that removing FaceNet results in degraded image quality with excessive smoothing and missing facial details (e.g., wrinkles). In the 4th row, removing Dlib produces noticeable visual flaws, such as artifacts in the teeth, color inconsistencies around the eyes and unnatural textures in wrinkles. In contrast, the full model delivers the best visual quality and preserves fine facial details.


\subsection{Downstream Applications}
Based on the face-swapped 3DGS avatar generated by our GaussianSwap, we implement three downstream applications. The visual results are presented in the supplementary video.
\subsubsection{Video Face Reenactment.}
This application can be viewed as the video-driven 3DGS avatar animation. Given a driven video, we conduct the FLAME tracking \cite{DECA2021} which has been described in the main body of the manuscript to obtain the camera poses and FLAME parameters. The camera poses, expression, head rotation and jaw pose parameters in all frames are transferred into the 3DGS avatar for animation.

\subsubsection{Speech-driven 3DGS avatar animation.}
In this application, we use the speech signals to generate the corresponding 3D face motions, which further drive the 3DGS avatar. We employ the speech-to-motion model proposed in Learn2Talk~\cite{Learn2Talk2025}. The sequence of mesh models generated by Learn2Talk is converted into FLAME parameters through 3D template fitting. These FLAME parameters are then input into the 3DGS avatar, producing the final 3DGS facial animation.

\subsubsection{Dynamic Background Manipulation.}
When projecting the 3DGS avatar into the image plane, we can obtain an alpha channel map for each frame. We then blend the 3DGS avatar with a new background using alpha compositing.

\section{Conclusion and Limitation}
In this paper, we introduce GaussianSwap, a novel video face swapping framework that constructs a 3DGS-based head avatar from a target video while transferring identity from a source image to the avatar. We believe the idea of empowering face manipulation through 3D head avatar reconstruction can inspire advancements in other face generation tasks.

As shown in the quantitative evaluation, our method doesn't show advantage in preserving facial expression and pose attributes from the target video. This is primarily due to the inherent drawback of FLAME tracking \cite{DECA2021}, which can't extract highly precise expression and pose parameters from the target video. The continuous progress in the field of 3D face tracking can help mitigate this issue, such as the application of $L_0$ optimization techniques~\cite{L0Xuli, L02014, L02016, PointsL0} for improved tracking accuracy.

Additionally, the constructed 3DGS avatar may
exhibit artifacts in side-view rendering if the target video lacks side-view perspectives. This limitation persists in both previous NeRF-based methods and concurrent 3DGS-based methods. Improving the generalization capability to handle dramatically novel views remains an open research challenge for 3DGS.


\bibliographystyle{IEEEtran}
\bibliography{tmm2026}

\end{document}